\documentclass[10pt,letterpaper]{article}
\usepackage{acl2012}
\usepackage{times}
\usepackage{url}
\usepackage[utf8x]{inputenc}
\usepackage[pdftex]{graphicx}
\usepackage{latexsym}
\usepackage{amsmath}
\usepackage{verbatim}

\title{Identifying Subjective and Figurative Language in Online Dialogue}

\author{Stephanie M. Lukin, Luke Eisenberg, Thomas Corcoran, \& Marilyn A. Walker\\
	    Natural Language and Dialogue Systems\\
		Computer Science Department, SOE-3\\
	    University of California, Santa Cruz\\
	    {\tt slukin,leisenbe,tcorcora,mawalker@ucsc.edu}}

\date{}

\begin{document}
\maketitle

More and more of the information on the web is dialogic, from Facebook
newsfeeds, to forum conversations, to comment threads on news
articles.  In contrast to traditional, monologic 
resources such as news,
highly social dialogue is very frequent in social media, as
illustrated in the snippets in Fig.~\ref{sample-quote-response} from
the publicly available Internet Argument Corpus ({\bf IAC})
\cite{Walkeretal12c}.  Utterances are frequently sarcastic, e.g., {\it
  Really?  Well, when I have a kid, I’ll be sure to just leave it in
  the woods, since it can apparently care for itself} (R2 in
Fig.~\ref{sample-quote-response} as well as Q1 and R1), and are often
nasty, (R2 in Fig.~\ref{sample-quote-response}).  Note also the
frequent use of dialogue specific discourse cues, e.g. the use of
{\it No} in R1, {\it Really? Well} in R2, and {\it okay, well} in Q3
in Fig.~\ref{sample-quote-response}~\cite{FoxTreeSchrock99,BryantFoxtree02,FoxTree10}.

\begin{figure}[ht!b]
\begin{center}
\begin{scriptsize}
\vspace{-.1in}
\begin{tabular}{|p{2.2in}|p{.2in}|p{.23in}|}
\hline  
Quote {\bf Q}, Response {\bf R} & {\bf Sarc} & {\bf Nasty} \\ \hline
{\bf Q1}: I jsut voted. sorry if some people actually have, you know, LIVES and don't sit around all day on debate forums to cater to some atheists posts that he thiks they should drop everything for. emoticon-rolleyes emoticon-rolleyes emoticon-rolleyes As to the rest of your post, well, from your attitude I can tell you are not Christian in the least. Therefore I am content in knowing where people that spew garbage like this will end up in the End. &  & \\
{\bf R1}: No, let me guess . . . er . . . McDonalds. No, Disneyland. Am I getting closer? & 1 & -3.6
\\ \hline  \hline 
{\bf Q2}: The key issue is that once children are born they are not physically dependent on a particular individual. &  &  \\
{\bf R2}  Really? Well, when I have a kid, I'll be sure to just leave it in the woods, since it can apparently care for itself. & 1 & -1 \\
\hline  
\end{tabular}
\end{scriptsize}
\end{center}
\vspace{-.1in}
\caption{\label{sample-quote-response} Sample Quote/Response Pairs
  from {\small \tt 4forums.com} with Mechanical Turk annotations for
  Sarcasm and Nasty/Nice. Highly negative values of Nasty/Nice
  indicate strong nastiness and sarcasm is indicated
by values near 1.}
\vspace{-.15in}
\end{figure}

\begin{figure}
\begin{center}
\vspace{-.1in}
\includegraphics[width=3.0in]{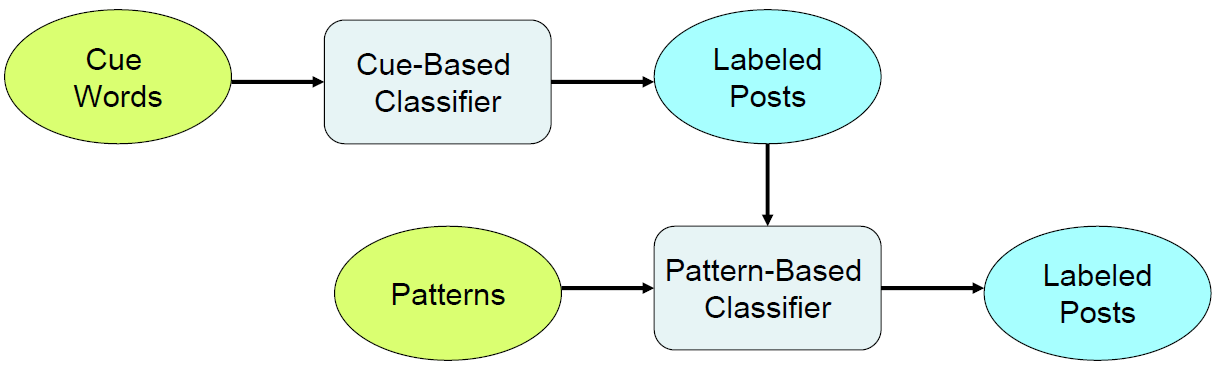}
\vspace{-.15in}
\caption{\label{rw} {\small Bootstrapping Method}}
\end{center}
\vspace{-.3in}
\end{figure}

We aim to automatically identify sarcastic and nasty utterances in unannotated online dialogue, extending a bootstrapping method previously applied to the classification of monologic subjective sentences by Riloff \& Wiebe, henceforth R\&W \cite{RiloffWiebe03,ThelenRiloff02}. We look at both sarcastic and nasty dialogic turns as a way to explore generalization of the method. R\&W's method creates a High-Precision, Cue-Based Classifier to be a first approximation on unannotated text. They improve their classifier by learning and bootstrapping patterns (Fig.~\ref{rw}).

We found that this bootstrapping method `as is' is not appropriate for our data because our Cue-Based Classifier yields a much lower precision than the bootstrapping requires. We have adapted the method to fit the sarcastic and nasty dialogic domain. Our method is as follows:
\vspace{-.05in}
\begin{enumerate}
\item Explore methods for identifying sarcastic and nasty cue words and phrases in dialogues; 
\vspace{-.1in}
\item Use the learned cues to train a sarcastic (nasty) Cue-Based Classifier 
\vspace{-.1in}
\item Learn general syntactic extraction patterns from the sarcastic (nasty) utterances and define fine-tuned sarcastic patterns to create a Pattern-Based Classifier;
\vspace{-.1in}
\item Combine both Cue-Based and fine-tuned Pattern-Based Classifiers to maximize precision at the expense of recall and test on unannotated utterances.
\vspace{-.2in}
\end{enumerate}

{\it Cue Words.} Sarcasm is known to be highly variable in form, and to depend, in some
cases, on context for its interpretation \cite{SW81,Gibbs00,BryantFoxtree02}. 
We elicit annotations from Mechanical Turk to identify
sarcastic (nasty) cues in utterances from 
a development set.
Turkers were presented with dialogic turns (a quote and its response) previously
labeled sarcastic or nasty in the IAC by 7 different annotators, 
and were asked to identify sarcastic (nasty) or potentially sarcastic (nasty) phrases 
in the turn response. The Turkers then selected words or phrases from the response 
they believed could lead someone to believing 
the utterance was sarcastic or nasty. 
\cite{Snowetal08} measure the quality of Mechanical Turk annotations on common 
NLP tasks by comparing them to 
a gold standard. Pearson's correlation coefficient shows that very few
Mechanical Turk annotators were required to beat the gold standard data, 
often less than 5. 
Because our sarcasm (nasty) task does not have gold standard data, we asked 
100 annotators to participate in the pilot. For all unigrams, bigrams, and trigrams, interannotator agreement plateaued around 20 annotators
and is about 90\%agreement with 10 annotators, showing that the Mechanical Turk
task is well formed and there is high agreement. 
We begin to form a sarcastic and nasty vocabulary from these cues.

{\it Cue based classifier.} We use a development set
 to measure ``goodness'' of a cue that could serve as a high
precision cue by using the percent sarcastic (nasty) and frequency
statistics in the development set. These features rely on 
how frequent ({\sc freq}) (subject to a $\theta_1$),
and how reliable ({\sc \%sarc} and {\sc \%nasty}) (subject to a $\theta_2$) a cue
has to be to be useful. We select candidate cues by exhausting $\theta_1$ $=$ [2, 4, 6, 8, 10] and $\theta_2$ $=$ [.55, .60,  .65, .70, .75, .80, .85, .90, .95, 1.00]  for
$\theta_1 \le $  {\sc freq} and $\theta_2 \le $ {\sc sarc}. At least two cues
must be present and above the thresholds in an utterance to be classified by the Cue-Based Classifier. Less than two cues are needed to be
classified as the counter-class. We select the best combination of 
parameters from our training set by selecting the parameters yielding the 
highest weighted f-measure that favors precision over recall. 
We then ran the Cue-Based Classifier with the best parameters on a test set. However as previously mentioned, R\&W's method expects the Cue-Based Classifier to yield high precision, whereas our results ({\sc cue} rows in Table~\ref{hp_pc_results}) are just barely above baseline. 

\begin{table}
\begin{scriptsize}
\begin{center}
\vspace{-0.2in}
\begin{tabular}{|p{0.55in}|r|c|c|c|c|}
\hline
SARC & PARAMS & P & R & F    \\ \hline \hline
Cue &  $\theta_1=$2, $\theta_2=$.55  & 51\% & 48\% & 0.5 \\ \hline
Baseline Pats &  $\theta_1=$2, $\theta_2=$.65  & 58\% & 78\% & 0.61 \\ \hline 
New Pats &  $\theta_1=$2,  $\theta_2=$.65 & 76\% & 79\% & 0.77 \\ \hline \hline
NASTY & PARAMS & P & R & F    \\ \hline \hline
Cue &  $\theta_1=$2, $\theta_2=$.6  & 66\% & 40\% & 0.58 \\ \hline
Baseline Pats &  $\theta_1=$2, $\theta_2=$.7  & 86\% & 55\% & 0.77 \\ \hline
New Pats &  $\theta_1=$2, $\theta_2=$.7  & 100\% & 5\% & 0.2 \\ \hline
\end{tabular}
\vspace{-.05in}
\caption{{\sc PARAMS}: the best parameters for 
each feature set P: precision, R: recall, F: weighted f-measure}
\vspace{-.3in}
\label{hp_pc_results} 
\end{center}
\end{scriptsize}
\end{table}

{\it Pattern Based Classifier.} The next step in R\&W's method is to create a Pattern-Based Classifier that takes as input the predicted labels from the Cue-Based Classifier. R\&W's Pattern-Based Classifier is trained on general, syntactic templates known to exist for subjectivity. These patterns are not limited to exact surface matches as the Cue-Based Classifiers require. We reimplement these patterns, and further 
developed new patterns specifically fine-tuned
towards sarcasm in dialogue. For example, our new pattern {\tt \small
  OH RB} (oh adverb) matches utterances like ``oh right" and ``oh sorry" and 
the pattern {\tt \small NP WHphrase} matches ``someone who" and ``someone what". 
Patterns are extracted from another development set and we
again compute {\sc freq} and {\sc \%sarc} and {\sc \%nasty} for each
pattern subject to $\theta_1 \le$ {\sc freq} and $\theta_2 \le$ {\sc
  \%sarc} or {\sc \% nasty}.  Classifications are made if at least two
patterns are present and both are above the specified $\theta_1$ and
$\theta_2$, again exhausting all combinations of $\theta_1$ and $\theta_2$.  Also following  R\&W, we do not learn
``not sarcastic" or ``nice" patterns. The counter-classes are predicted when the
utterance contains less than two patterns. We test two Pattern-Based Classifiers: one with the original patterns proposed in R\&W ({\sc baseline pats}) and one with the original patterns in addition to our new, fine-tuned patterns ({\sc new pats}). Table~\ref{hp_pc_results} shows the results of the parameters with the highest weighted f-measure. 

The Pattern-Based Classifier performs better on Nasty than Sarcasm. We conclude that R\&W's patterns alone generalize well on our Sarcasm and Nasty datasets. By adding the fine-tuned patterns in the {\sc new pats} Classifier, we see a drastic increase in Sarcasm precision. There seems to be little change in recall for Sarcasm. Furthermore, we see a huge increase in precision for Nasty, but a steep decline in recall with the new patterns. We believe this is because these new patterns are tailored towards sarcastic utterances, not nasty. We did not create our own fine-tuned nasty patterns because we do well with R\&W's general patterns. 

\begin{table}
\begin{scriptsize}
\begin{center}
\vspace{-0.2in}
\begin{tabular}{|c|c|c|c|}
\hline
sarcasm					& P & R & F    \\ \hline \hline
cue-based			&  51\% & 48\% & 0.5 \\ \hline
cue {\sc or} patterns 	&  56\% & 62\% & 0.57 \\ \hline
cue {\sc and} patterns 	&  71\% & 32\% & 0.57 \\ \hline \hline
nasty					& P & R & F    \\ \hline \hline
cue-based			&  66\% & 40\% & 0.58 \\ \hline
cue {\sc or} patterns 	&  75\% & 44\% & 0.69 \\ \hline 
cue {\sc and} patterns 	&  88\% & 31\% & 0.44 \\ \hline

\end{tabular}
\vspace{-.05in}
\caption{\label{combined}; Compares the Cue-Based Classifier to the Combined Classifier;
P: precision, R: recall, F: f-measure}
\vspace{-.3in}
\end{center}
\end{scriptsize}
\end{table}

{\it Combined Classifier.} To attempt to create a High-Precision Classifier, we combine the Cue-Based Classifier and the Pattern-Based Classifier. We classify a post as sarcastic if it meets either the criteria of the Cue-Based Classifier (e.g. with $\theta_1=2, \theta_2=.55$ for Sarcasm) or the Pattern-Based Classifier (e.g. with $\theta_1=2, \theta_2=.65$ for Sarcasm). We use the same test set with which we test the Cue-Based Classifier and compare the results (Table~\ref{combined}). We furthermore distinguish between a Combined Classifier that makes a classification if both schemata are true ({\sc and}), or if only one is true ({\sc or}).

{\sc or} does better than only the Cue-Based Classifier for all precision, recall, and f-measure. 
{\sc and} does better for precision by far than the Cue-Based Classifier, but with a lower recall. Despite the very low recall for {\sc and}, the f-measure of {\sc and} and {\sc or} is identical. {\sc and} is a more selective classifier, only saying ``yes" if both schemata are true. This will naturally yield lower recall, but grant higher confidence in those classified. 

We believe our Combined {\sc and} Classifier now has a high enough precision to be compared with R\&W's first approximation High-Precision, Cue-Based Classifier. After running the Combined Classifier on unannotated data, we select 100 predicted sarcastic and 100 predicted not sarcastic utterances and ask human annotators to label them. We expect a high overlap between annotators and the Combined Classifier, which would indicate that human annotators agree with the labels we are automatically predicting. These results are currently in progress. 

Despite the fact that we could not create a first approximation High-Precision, Cue-Based classifier like R\&W, we have succeeded in creating a High-Precision Combined Classifier using both cues and fine-tuned patterns (71\% precision for sarcasm and 88\% precision for nastiness). Future work will involve developing fine-tuned patterns for nastiness and exploring different patterns for sarcasm.

\pagebreak

\bibliographystyle{naaclhlt2013}
\bibliography{nl}

\end{document}